\documentclass[iicol,sn-mathphys-num, pdflatex]{sn-jnl}

\usepackage{graphicx}%
\usepackage{multirow}%
\usepackage{amsmath,amssymb,amsfonts}%
\usepackage{amsthm}%
\usepackage{mathrsfs}%
\usepackage[title]{appendix}%
\usepackage{xcolor}%
\usepackage{textcomp}%
\usepackage{manyfoot}%
\usepackage{booktabs}%
\usepackage{makecell}
\usepackage{listings}%
\makeatletter  
\newif\if@restonecol
\def\BState{\State\hskip-\ALG@thistlm}
\makeatother

\usepackage[linesnumbered,ruled,vlined]{algorithm2e}
\usepackage{algpseudocode}  
\usepackage{amsmath}  

\theoremstyle{thmstyleone}%
\theoremstyle{thmstyletwo}%

\theoremstyle{thmstylethree}%

\begin{document}

\title[Fusion of Multi-scale Heterogeneous Pathology Foundation Models for Whole Slide Image Analysis]{Fusion of Multi-scale Heterogeneous Pathology Foundation Models for Whole Slide Image Analysis}







\author[1,2,3]{Zhidong {Yang}}
\equalcont{\small These authors contributed equally to this work.}
\author[4]{Xiuhui {Shi}}
\equalcont{\small These authors contributed equally to this work.}
\author[10]{Wei {Ba}}
\author[10]{Zhigang {Song}}
\author[6]{Haijing {Luan}}
\author[6]{Taiyuan {Hu}}
\author[7]{Senlin {Lin}}

\author*[2,3]{Jiguang {Wang}}
\email{jgwang@ust.hk}
\author*[1,5,8,9]{Shaohua Kevin {Zhou}}
\email{skevinzhou@ustc.edu.cn}
\author*[1,5]{Rui {Yan}}
\email{yanrui@ustc.edu.cn}

\affil[1]{\small School of Biomedical Engineering, Division of Life Sciences and Medicine, University of Science and Technology of China, Hefei, Anhui, China}
\affil[2]{\small Division of Life Science, Department of Chemical and Biological Engineering, State Key Laboratory of Nervous System Disorders, The Hong Kong University of Science and Technology, Hong Kong SAR, China}
\affil[3]{\small SIAT-HKUST Joint Laboratory of Cell Evolution and Digital Health, HKUST Shenzhen-Hong Kong Collaborative Innovation Research Institute, Futian, Shenzhen, China}
\affil[4]{\small Department of Hepatobiliary Surgery, The First Affiliated Hospital of USTC, Division of Life Sciences and Medicine, University of Science and Technology of China, Hefei, Anhui, China}
\affil[5]{\small Center for Medical Imaging, Robotics, Analytic Computing \& Learning (MIRACLE), Suzhou Institute for Advanced Research, USTC, Suzhou, Jiangsu, China}
\affil[6]{\small Computer Network Information Center, Chinese Academy of Sciences, Beijing, China}
\affil[7]{\small Institute of Computing Technology, Chinese Academy of Sciences, Beijing, China}
\affil[8]{\small Jiangsu Provincial Key Laboratory of Multimodal Digital Twin Technology, Suzhou, Jiangsu, China}
\affil[9]{\small Key Laboratory of Precision and Intelligent Chemistry, USTC, Hefei, Anhui, China}
\affil[10]{\small Department of Pathology, Chinese PLA General Hospital, Beijing, China}

\abstract{Whole slide image (WSI) analysis has emerged as an increasingly essential technique in computational pathology. Recent advances in the pathology foundation models (FMs) have demonstrated significant advantages in deriving meaningful patch-level or slide-level multi-scale features from WSIs.
However, current pathology FMs have exhibited substantial heterogeneity caused by diverse private training datasets and different network architectures. This heterogeneity introduces performance variability when we utilize the features from different FMs in the downstream tasks. To fully explore the advantages of multiple FMs effectively, 
in this work, we propose a novel framework for the fusion of multi-scale heterogeneous pathology FMs, called FuseCPath, yielding a model with a superior ensemble performance. The main contributions of our framework can be summarized as follows: 
\textbf{(i)} To guarantee the representativeness of the training patches, we propose a multi-view clustering-based method to filter out the discriminative patches via multiple FMs' embeddings. \textbf{(ii)} To effectively fuse the patch-level FMs, we devise a cluster-level re-embedding strategy to online capture patch-level local features. \textbf{(iii)} To effectively fuse the slide-level FMs, we devise a collaborative distillation strategy to explore the connections between slide-level FMs. 
Extensive experiments demonstrate that the proposed FuseCPath achieves state-of-the-art performance across multiple tasks on diverse datasets.}





\maketitle

\section{Introduction}\label{sec1}
\begin{figure*}[!h]
\centerline{\includegraphics[width=\textwidth]{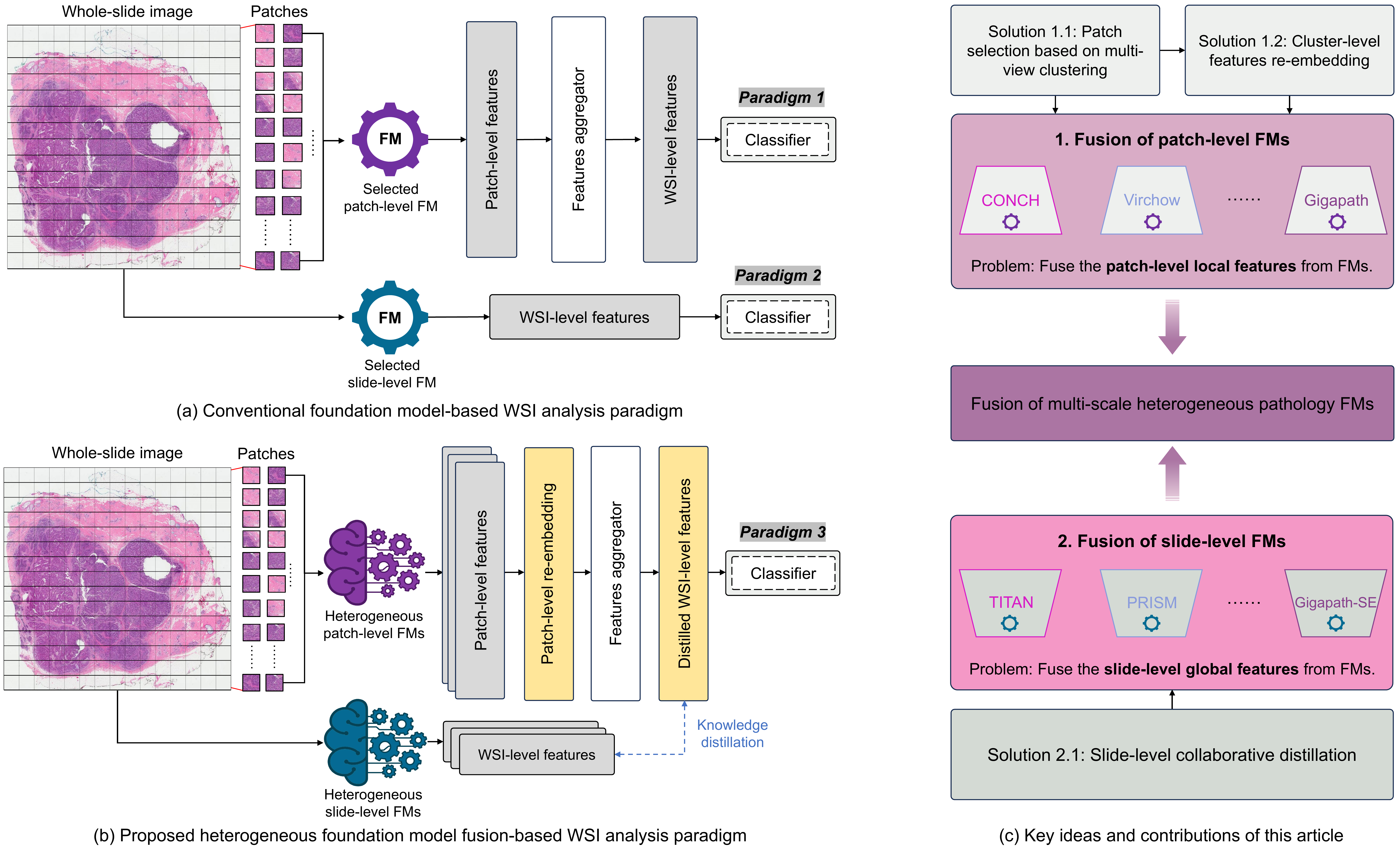}}
\caption{(a) Conventional foundation model-based WSI analysis paradigm. To achieve optimal performance on downstream tasks, the most straightforward strategy is to select a patch-level or slide-level foundation model that exhibits the strongest performance on the target task. (b) Proposed heterogeneous foundation model fusion-based WSI analysis paradigm. Based on the concept of ensemble learning, a framework for the fusion of multi-scale (patch-level and slide-level) heterogeneous pathology FMs will yield a model with superior performance. (c) The key ideas and contributions of this article.}
\label{fig1}
\end{figure*}
Pathological diagnosis is the gold standard for cancer diagnosis, while whole slide image (WSI) analysis occupies a core position in computational pathology (CPath) and can support key tasks such as cancer subtyping \citep{LU2022102298,huang2024free}, survival prediction \citep{10226279,jaume2023modeling,yan2025tpami}, and biomarker prediction \citep{YAN2023102824,cobra,suretrans}. In recent years, the rapid development of pathology foundation models (FMs) \cite{conch,virchow2,prism,chief} has brought about a revolutionary transformation in this field.

Current pathology FMs can be categorized into two distinct types at different scales, which are patch-level FMs \citep{conch,uni,virchow2} and slide-level FMs \citep{gigapath,prism,titan,chief}. The patch-level FMs are trained with the tiled patches of WSIs. The patch embeddings derived from patch-level FMs will be aggregated with Multiple-Instance Learning (MIL) for the training of downstream tasks in CPath. Most of the patch-level FMs are trained with self-supervised learning methods of different architectures (e.g., Dino-v2 \citep{dinov} or MAE \citep{mae}), using different private datasets. Different from patch-level FMs, the slide-level FMs are capable of constructing slide embeddings with unsupervised learning. Similar to patch-level FMs, the architectures of backbone models and training datasets differ significantly in each slide-level FM. In conclusion, we define these FMs differences as \textit{heterogeneity} in the pathology FMs.

Because of the heterogeneity, the performance in different downstream tasks and the learned tissue morphologies are diverse across different FMs \citep{benchcpath}. To ensure the performance of FMs on downstream tasks, the most trivial strategy is to select a foundation model with the best performance on the corresponding downstream task, as shown in Figure \ref{fig1}(a). However, this strategy contains obvious shortcomings. Firstly, re-training a foundation model with our own training datasets may not reproduce the optimal performance. Secondly, when we are facing more than one downstream task, re-training many FMs simultaneously is not a flexible solution. Consequently, based on the concept of ensemble learning, it is an effective way to fuse the patch-level and slide-level embeddings from the heterogeneous FMs into a single proxy model for training, as shown in Figure \ref{fig1}(b). By combining the strengths of each individual foundation model, we will obtain a fused model with improved performance on downstream tasks \citep{benchcpath}. However, there still exist two major challenges hindering the fusion of heterogeneous FMs. Firstly, the heterogeneity in the pathology FMs contributes to diverse dimensions and information of the embeddings. It is essential to comprehensively capture the connections between the patch-level or slide-level embeddings derived from heterogeneous FMs. Secondly, the scale gaps between patch-level and slide-level embeddings. The representation information captured by patch-level and slide-level embeddings is distributed at different scales. We need to fully leverage the representational information from slide-level embeddings to assist in training models with patch-level embeddings.

To address these challenges, in this work, we propose a novel framework called FuseCPath for the fusion of multi-scale (patch-level and slide-level) heterogeneous pathology FMs. Figure \ref{fig1}(c) illustrates the key ideas and contributions of the proposed FuseCPath. The fusion of patch-level FMs aims at fusing the patch-level local features from diverse FMs (Figure \ref{fig1}(c)). First, we propose a multi-view clustering strategy to select representative patches utilizing the meaningful features captured from heterogeneous patch-level FMs. Second, we devise a cluster-level feature re-embedding transformer to discover the relations between patch-level FMs in feature space. The fusion of slide-level FMs aims at fusing the slide-level global features from diverse FMs (Figure \ref{fig1}(c)). Consequently, we devise a collaborative distillation module to effectively utilize the representative global features residing in slide embeddings as a teacher model. Equipped with the above modules, the FuseCPath framework will be capable of fusing the heterogeneous FMs effectively. The code is publicly available from \url{https://github.com/ZhidongYang/FuseCPath}.

\begin{itemize}
\item We propose FuseCPath, a novel framework for fusing the heterogeneous multi-scale (patch-level and slide-level) pathology FMs to integrate a model with better performance.

\item We devise a novel online feature re-embedding transformer that operates on filtered discriminative patch-level features with multi-view clustering. The proposed online re-embedding effectively addresses the issue of fusing the heterogeneous patch-level FMs by capturing meaningful features locally and connecting the patch embeddings across diverse patch-level FMs.

\item We propose a novel collaborative distillation module for the fusion of slide-level FMs that systematically bridges the scale gap between the fusion of patch-level and slide-level FMs. The slide-level FMs will serve as teacher models to provide global representations of WSIs.

\item Extensive experiments demonstrate that the FuseCPath will ensemble a new model with superior performance across various downstream tasks in WSI analysis.
\end{itemize}

\section{Related work}

\subsection{Pathology foundation model}
Recent advances in pathology FMs have employed diverse architectural and training paradigms, predominantly utilizing self-supervised learning (SSL) techniques \citep{byol2020,simclr} to extract meaningful representations from unannotated patches in WSIs. SSL-based approaches can be summarized as follows: (1) contrastive learning frameworks such as REMEDIS \citep{remedis}, which adapt SimCLR frameworks \citep{simclr} by maximizing feature similarity between comparable regions within individual WSIs while minimizing the similarity across disparate slide regions; (2) masked image modeling adopted by Prov-gigapath \citep{gigapath}, CONCH \citep{conch}, and BEPH \citep{beph}, where random patch occlusion forces models to reconstruct masked tissue patterns, thereby capturing robust contextual relationships; and (3) knowledge distillation implementations exemplified by Virchow2 \citep{virchow2}, UNI \citep{uni}, and Hibou \citep{hibou}, which employ DINO-based frameworks \citep{dinov} to distill the knowledge from teacher to student models, yielding compact yet generalizable representations without extensive annotations.

Slide-level representation learning has emerged as an essential approach for generating task-agnostic embeddings through unsupervised learning. Pioneering work by \citet{Chen2022HIPT} proposed the HIPT method by devising a hierarchical self-distillation for WSI-level representation learning. \citet{gigassl} developed a contrastive learning-based framework using augmented patch ensembles. Subsequent innovations include Prov-gigaPath (SE)'s masked autoencoder architecture \citep{gigapath} for generating slide representations and several multi-modal-based pretraining FMs \citep{titan,prism}. Existing slide-level FMs universally require substantial training data (over 10K WSIs) \citep{Chen2022HIPT,gigassl,chief}, while PRISM \citep{prism} and GigaPath-SE (slide encoder) \citep{gigapath} utilize more WSIs. With the rapid development of multi-omics techniques, FMs will be capable of bridging the H\&E-stained pathology images to other omics data \citep{threads, madeline, omiclip}.

\subsection{Multiple instance learning in WSI analysis}

Multi-instance learning (MIL) is a predominant weakly supervised learning strategy widely adopted in the applications of downstream tasks for WSI analysis \citep{wang2019rmdl,lu2021data,li2021dual}, which solves the problem of lacking high-quality annotations. The attention-based deep multi-instance learning (AB-MIL) proposed by \cite{ilse2018attention} first adopts convolutional neural networks (CNNs) to multi-instance learning. This technique is widely extended to the application for WSI image analysis. \cite{wang2019rmdl} introduced a recalibrated multi-instance learning framework (RMDL) for the classification of whole slide images (WSIs) of gastric tissues. The RMDL employs a convolutional neural network (CNN) to identify discriminative instances (the image patches) within each WSI and subsequently trains the model exclusively on these selected instances. RMDL captures dependencies among instances and dynamically recalibrates their features based on the coefficients derived from fused feature representations. \cite{li2021dual} developed a dual-stream multiple instance learning network (DSMIL) comprising two synergistic streams: one learns an instance-level classifier using max-pooling to identify the highest-scoring (critical) instance, while the other computes attention scores for instances based on their proximity to the critical instance. \cite{yao2020whole} proposed DeepAttnMISL, a survival prediction model that integrates attention mechanisms with multi-instance learning. This approach clusters the patches extracted from WSIs into phenotypically distinct groups. Then it selects representative patches from each cluster and processes them through a Siamese multi-instance fully convolutional network. The model subsequently aggregates features via attention-based multiple instance learning (AB-MIL) pooling to predict patient survival risk. Similarly, \cite{lu2021data} proposed CLAM, which also operates in two stages: first, patches are encoded into feature vectors using a pre-trained CNN, and then these features are processed by a clustering-constrained attention mechanism within a multiple instance learning framework to produce final predictions. \cite{schirris2022deepsmile} developed DeepSMILE, a two-stage framework wherein the first stage employs the contrastive learning method SimCLR for patch-level feature extraction, generating representative feature embeddings. The second stage incorporates these features into the proposed VarMIL, which is an extension of AB-MIL that introduces a feature variability module to explicitly model tumor heterogeneity. \citet{YAN2023102824} proposed a hierarchical deep multi-instance learning-based framework called HD-MIL to accurately predict gene mutations in bladder cancer by leveraging a contrastive learning framework called Bootstrap Your Own Latent (BYOL) to derive high-quality feature representations. \citet{MambaMIL} proposed to incorporate the Selective Scan Space State Sequential Model (Mamba) in Multiple Instance Learning (MIL) for long sequence modeling with linear complexity to adjust the high-resolution of WSIs. Similarly, \citet{tang2024feature} proposed a re-embedding strategy called R$^2$T to online captures foundation model-level local features and establishes connections across different regions. Additionally, the proposed R$^2$T can be integrated into MIL models (R$^2$T-MIL) to improve the performance of several downstream tasks.

\subsection{Patch selection in WSI analysis}
Due to the gigapixel-scale high resolution of WSIs, it is challenging to fit WSIs to the GPU devices in an end-to-end manner. One effective solution is to crop the images into patches for training. Several approaches are proposed to implement this module. \cite{li2018graph} first proposed DeepGraphSurv, a survival analysis model that employs graph convolutional networks (GCNs) by randomly sampling over 1,000 patches from each WSI to construct graphs for classification, achieving C-indices of 0.66 and 0.62 on TCGA-LUSC and TCGA-GBM datasets, respectively. \cite{raju2020graph} proposed an integrated framework combining graph neural networks with attention-based multiple instance learning for colorectal cancer TNM staging, where they extracted texture features from randomly selected patches, constructed graphs from these patches, and used them as instances in their classification model. While demonstrating broad applicability and straightforward implementation, this approach may be limited by the potential lack of representativeness in randomly sampled patches, which could impact classification performance.

To ensure the representativeness of patches, the strategy of approximating Regions of interest (RoI) is adopted. The RoI can be approximated using several distinctive patches, and several notable methods have been developed based on this conclusion (\cite{hou2016patch,adnan2020representation,wang2019weakly}). For instance, \cite{adnan2020representation} employed the color-based strategy outlined in Yottixel (\cite{kalra2020yottixel}) to extract several patches from WSIs, and these patches are modeled by a fully connected graph. In this way, the task of classifying WSIs is converted into graph classification. In \cite{adnan2020representation}, the authors gathered 1,026 WSIs from the TCGA lung cancer dataset, achieving an accuracy of 88.8. \cite{wang2019weakly} utilized weakly supervised learning to categorize lung cancer into four subtypes. This method first utilizes a patch-based full convolutional neural network to identify distinctive blocks, then applies different block selection and feature aggregation strategies based on probability maps to generate a global representation for the WSI. Finally, these global representations will serve as input to a random forest, which will produce the classification results.

The clustering strategy is also an effective way to provide prior knowledge for patch selection. Based on the result of the feature clustering, several clustering-based methods (\cite{zhu2017wsisa,xu2022spatial,YAN2023102824}) are proposed to guarantee the representativeness of the selected patches. The survival prediction method (WSISA) developed by \cite{zhu2017wsisa} can make effective use of all distinguishing patch features in WSIs, thereby significantly enhancing survival prediction performance compared to existing methods. WSISA first selects hundreds of patches from each WSI and then further clusters these selected patches. Then, it selects clusters based on the patch-level prediction performance using CNN, and fuses them to make the final prediction. Inspired by this method, \cite{xu2022spatial} firstly combines the advantages of selecting patches in Regions-of-Interest (RoI) from detected cancer areas and clusters, using the data from 253 bladder cancer patients in the TCGA dataset. The method proposed by  \cite{YAN2023102824} proposed to select representative patches from clustered detected cancer areas using high-quality embeddings derived from a BYOL-based pre-trained model.

\section{Method}
\begin{figure*}[!htbp]
\centerline{\includegraphics[width=0.9\textwidth]{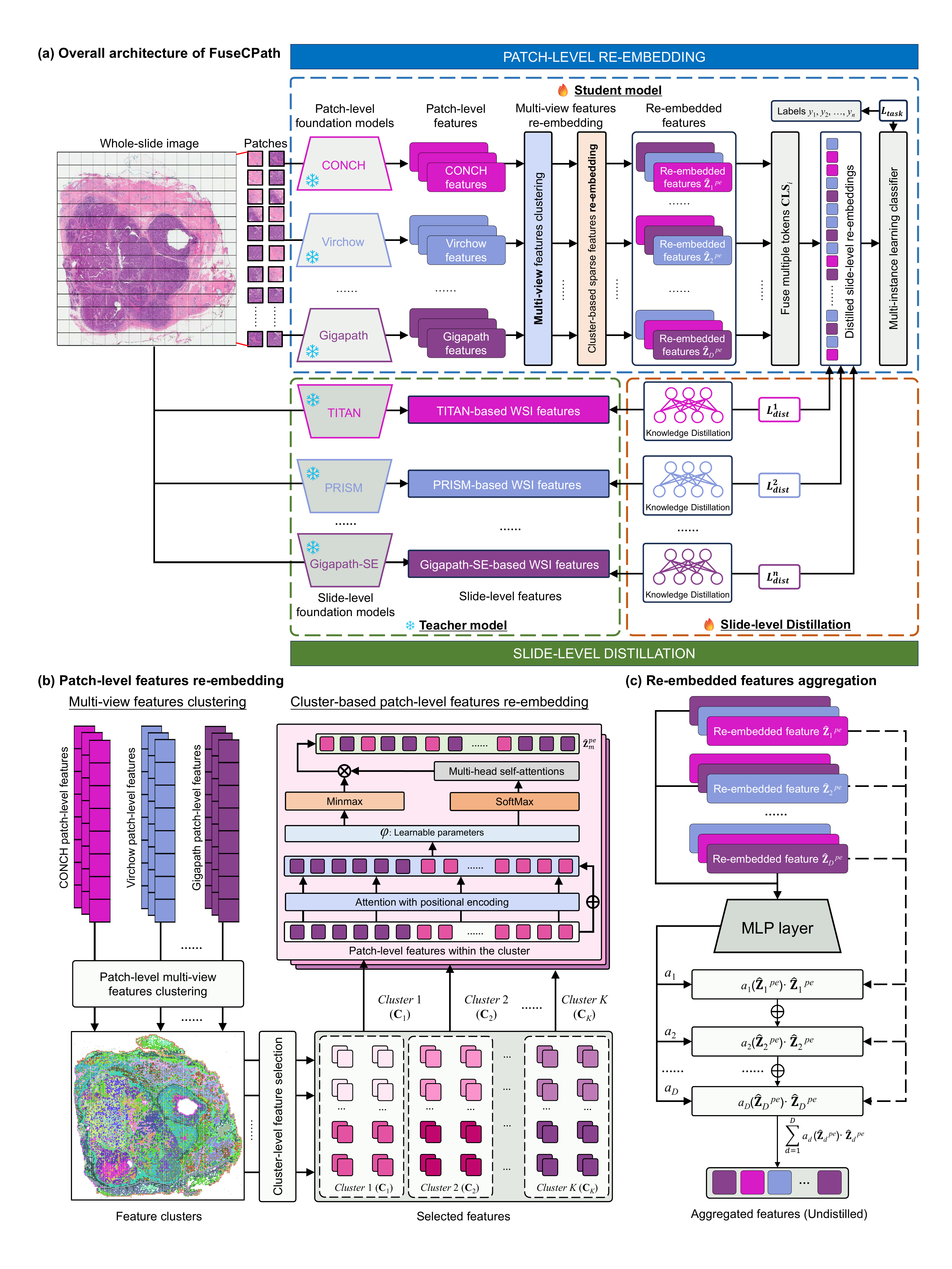}}
\caption{The overall architecture and main components of the proposed FuseCPath framework. (a) The overall architecture of the FuseCPath framework. The FuseCPath can be divided into two essential branches, which are patch-level features re-embedding and slide-level features collaborative distillation. (b) The demonstration of patch-level features re-embedding. Representative features can be summarized with multi-view clustering and online re-embedding. (c) The re-embedded features aggregation module is implemented by AB-MIL. }
\label{fusecpath}
\end{figure*}

FuseCPath is a framework for the fusion of multi-scale (patch-level and slide-level) heterogeneous pathology FMs, which contributes to a significant performance improvement on the WSI image analysis across multiple tasks. We provide a brief overview of FuseCPath below, and Figure \ref{fusecpath}(a) presents more details. Given a set of WSIs $\{X_i | X_i\in \mathbb{R}^{d_x\times d_y\times 3}\}$, FuseCPath will simultaneously fuse the patch embeddings and slide embeddings of $X_i$. For the fusion of patch-level FMs, the FuseCPath will first cluster the patch embeddings with multiview spectral clustering to find representative patches.  Then, a Cluster-level Re-embedding Transformer (CR$^2$T) is used to online fuse the patch embeddings, and the Attention-Based Multiple instance learning (AB-MIL) to aggregate the re-embedded features. For the fusion of slide-level FMs, FuseCPath regards the slide-level embeddings as the teacher models' information, which is implemented with a collaborative distillation during training.

\subsection{Multi-view patch features clustering}
\begin{figure}[!t]
\centerline{\includegraphics[width=\columnwidth]{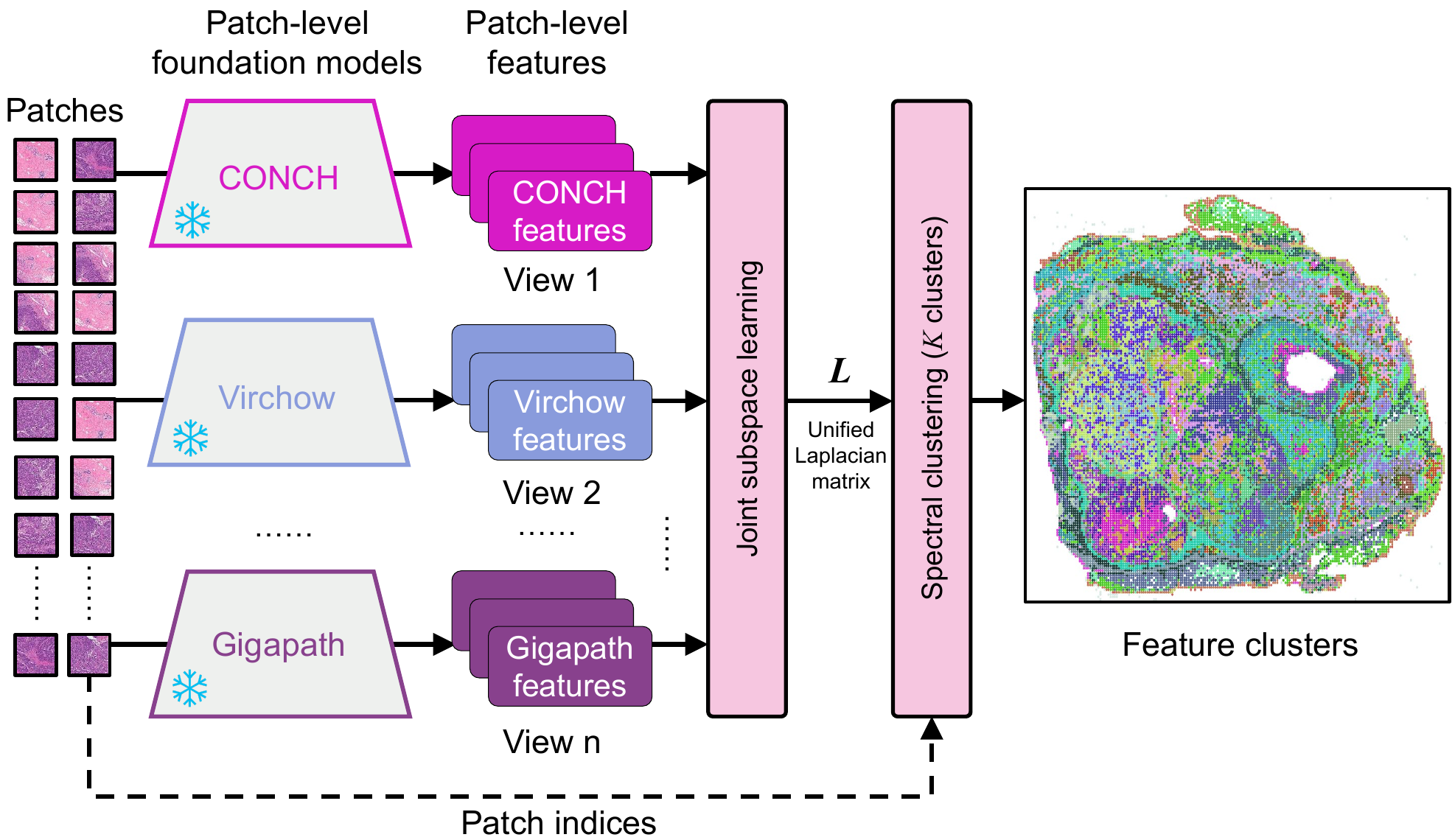}}
\caption{Details of multi-view spectral clustering (MVSC). The patch embeddings from diverse patch-level FMs can be regarded as a view of the original WSI dataset.}
\label{fig3}
\end{figure}
Due to the extremely high resolution of the WSIs, it will be a challenging operation to input all the patches extracted from the WSIs for training. Consequently, a typical solution is to select a subset of the patches randomly with a fixed amount of patches. However, the random selection can not guarantee the representativeness of the patches for training. In this work, we devise a cluster-based strategy to select representative patch embeddings from multiple FMs for training.

Firstly, the patch embeddings are derived from pre-trained heterogeneous patch-level FMs.
\begin{equation}
\mathbf{H}^{f^{i}_{pe}}=f^{i}_{pe}(X),\mathbf{H}^{f^{i}_{pe}} \in \mathbb{R}^{N_X\times d^{i}_{pe}},
\label{pe}
\end{equation}
where $f^{i}_{pe}$ denotes the patch-level foundation model, and $f^{i}_{pe}\in \mathcal{F}_{pe}=$ \{CONCH, Virchow2, Gigapath\}. $\mathbf{H}^{f^{i}_{pe}}$ denotes the patch embeddings derived from the foundation model $\mathbf{H}^{f^{i}_{pe}}$. $N_X$ denotes the complete number of patches extracted from WSI $X$. $d^{i}_{pe}$ denotes the dimension of the embeddings extracted from $f^{i}_{pe}$, where $d^{i}_{pe}\in \mathcal{D}_{pe}=\{768,2560,1536\}$.

Since we need to simultaneously consider the representativeness of the selected patches based on the patch embeddings from multiple FMs $f^{i}_{pe}$, the traditional $K$-means cluster method is not suitable in this situation. Thus, the multi-view spectral clustering is selected as an optimal solution. The patch embeddings $\mathbf{H}^{f^{i}_{pe}}$ derived from a distinct foundation model can be regarded as a view of the original WSI, and each view provides a diverse representation of the WSI. 
\begin{equation}
\mathcal{H}^{f_{pe}}=\{\mathbf{H}^{f^{1}_{pe}},...,\mathbf{H}^{f^{n}_{pe}}\}.
\label{pelist}
\end{equation}
where the extended tensor $\mathcal{H}^{f_{pe}}$ is the input of multi-view spectral clustering. Consequently, the patches will be clustered into $K$ clusters, and then $N_K$ patches will be selected from the clusters. As a result, $N_K\times K$ patches will be selected as the representative patch for training embeddings. Figure \ref{fig3} illustrates the main process of multi-view clustering for heterogeneous patch embeddings from multiple patch-level FMs.

\subsection{Patch-level features re-embedding}

With the selected patch embeddings from the foundation model, existing methods chose to fine-tune the original model using the obtained features to adjust the downstream task. However, when the patch embeddings are derived from FMs with heterogeneous architectures, the models will be fine-tuned separately with our training data. From the perspective of MIL, this process can be formulated as follows.
\begin{equation}
z=\mathcal{A}(f^1_{pe}(X),...,f^n_{pe}(X)),
\label{finetune}
\end{equation}
where $z$ denotes the aggregated slide-level features using patch-level features from multiple sources of FMs $f^i_{pe}$. $\mathcal{A}(\cdot)$ denotes the mapping function of feature aggregation. The performance of this strategy is limited by the difference between our own fine-tuning datasets and the original training datasets for the FMs. A more effective solution is based on online simultaneous training using a consistent training paradigm. Thus, we devise an online cluster-level patch features re-embedding strategy to fuse the embeddings from heterogeneous FMs into a single model. This strategy can be formulated as follows.
\begin{equation}
z=\mathcal{A}(\mathcal{R}(f^1_{pe}(X),...,f^n_{pe}(X))),
\label{mre}
\end{equation}
where $\mathcal{R}(\cdot)$ denotes the mapping function of online features re-embedding. Inspired by R$^2$Transformer in \citep{tang2024feature}, we opt for the Regional Multi-head Self-attention (R-MSA) and Cross Regional Multi-head Self-Attention (CR-MSA) as the base model for our strategy. Figure \ref{fusecpath}(b) demonstrates the procedure of online re-embedding for patch-level features in our FuseCPath.

The number of patches in a high-resolution WSI is too large to serve as the input of the Transformer-based models. Especially in the situation of fusing the embeddings from multiple FMs, the embeddings are equipped with much higher dimensions. Thus, we need to avoid the Out-of-Memory issue. The R-MSA \citep{tang2024feature} strategy addresses this issue by partitioning the patches into independent regions, where multi-head self-attention is performed on these regions. In this work, the solution to this issue goes one step further. Unlike the vanilla R-MSA, the input embeddings for our re-embedding module have been summarized by multi-view clustering. Our R-MSA only needs to focus on the sparsely selected patches from the clusters, which are representative enough for training. Hence, this strategy can be refactored to \textbf{Clustered Multi-head Self-attention} (C-MSA). Previous work \citep{10226279} has proved the representativeness of the patches selected from the clusters. The C-MSA can be formulated as follows.
\begin{equation}
\begin{aligned}
& \mathbf{C}_1,...,\mathbf{C}_K=\text{Cluster}(\mathbf{H}^{f^{1}_{pe}},...,\mathbf{H}^{f^{n}_{pe}}), \\
& \mathbf{H}^{f^{1}_{pe}}_s,...,\mathbf{H}^{f^{N}_{pe}}_s=\text{Selection}(\mathbf{C}_1,...,\mathbf{C}_K), \\
& \mathbf{H}^{f^{i}_{pe}}_s\in \mathbb{R}^{(N_K\times K)\times d^{i}_{pe}}, \\
& \hat{\mathbf{Z}}^{pe}_m = \text{MSA}(\text{LN}([\mathbf{H}^{f^{1}_{pe}}_s,...,\mathbf{H}^{f^{N}_{pe}}_s]))+[\mathbf{H}^{f^{1}_{pe}}_s,...,\mathbf{H}^{f^{N}_{pe}}_s], \\
& \hat{\mathbf{Z}}_m^{pe}\in \mathbb{R}^{(N_K\times K)\times D},
\end{aligned}
\label{cluster}
\end{equation}
where $\mathbf{C}_1,...,\mathbf{C}_K$ denote the clusters. $\mathbf{H}^{f^{1}_{pe}}_s,...,\mathbf{H}^{f^{N}_{pe}}_s$ denote the selected patch embeddings from the clusters. $\hat{\mathbf{Z}}^{pe}_m$ denotes the encoded embeddings ($D=\Sigma_i d^{i}_{pe}$). We adopt the Position Encoding Generator (PEG) implemented using a 1-D convolutional layer to the encoded embeddings $\hat{\mathbf{Z}}^{pe}_m$.
\begin{equation}
\alpha_{ij}=\text{SoftMax}(\mathbf{e}_{ij}+\text{PEG}(\mathbf{e}_{ij})),
\label{peh}
\end{equation}
where $\alpha_{ij}$ is the attention weights of $\hat{\mathbf{Z}}^{pe}_{mj}$ with respect to $\hat{\mathbf{Z}}^{pe}_{mi}$. $\mathbf{e}_{ij}$ is a tensor calculated with a scaled dot-product attention using $\hat{\mathbf{Z}}^{pe}_m$ \citep{tang2024feature}.

Similar to Cross-regional Multi-Head Self-Attention (CR-MSA), it is essential to consider the semantic context between the selected patches for the downstream tasks in WSI analysis. Therefore, we need to model the connections between cluster-level patches using CR-MSA, which should be referred to as \textbf{Cross-cluster Multi-Head Self-Attention} (CC-MSA) in our work. The cluster-level features will be fused with the vanilla MSA module and normalized by the MinMax$(\cdot)$ function. This process can be formulated as follows.
\begin{equation}
\begin{aligned}
& \mathbf{R}^{pe}_a = \text{SoftMax}^{K}_{k=1}(\hat{\mathbf{Z}}^{pe}_{mk}\Phi)^T\hat{\mathbf{Z}}^{pe}_{m},\\
& \mathbf{W}^{pe}_d = \text{MinMax}^{K}_{k=1}(\hat{\mathbf{Z}}^{pe}_{mk} \Phi), \\
& \hat{\mathbf{W}}^{pe}_d=\text{SoftMax}^{G}_{g=1}(\hat{\mathbf{Z}}^{pe}_{mg}\Phi)\in \mathbb{R}^{G\times 1},\\
& \hat{\mathbf{Z}}^{pe} = (\mathbf{W}^{pe}_d)^T\text{MSA}(\mathbf{R}^{pe}_a)\hat{\mathbf{W}}^{pe}_d. \\
\end{aligned}
\label{ccmsa}
\end{equation}
where $\Phi\in \mathbb{R}^{D\times G}$ denotes learnable parameters, $\mathbf{W}^{pe}_d$ denotes the normalization weights for the fused patch-level features $\text{MSA}(\mathbf{R}^{pe}_a)$. The CC-MSA is calculated at the cluster level using patch embeddings.

\subsection{Re-embedded features aggregation}
Re-embedded features aggregation is an essential module of our heterogeneous FMs fusion framework. The multi-instance learning (MIL) is widely adopted as an effective solution in WSI analysis, where the labels $\mathbf{y}$ are assigned to each WSI. And the WSI $X$ can be defined as \textit{bag}, the patches within $X$ are \textit{instances}. To effectively demonstrate the features aggregation, we will first briefly introduce MIL and then proceed to the features aggregation in our FuseCPath.

The multi-instance learning (MIL) is a useful weakly supervised learning method in WSI analysis. The formulation of MIL is shown as follows. The dataset consists of bags assigned with labels $\mathbf{y}=\{y_1,y_2,...,y_D\}$. Each bag contains several instances. If at least one instance in a bag is positive, the bag will be considered positive, and if all instances in a bag are negative, the bag will be considered negative. Take the situation in binary classification as an example, we define $B = \left \{(x_1, y_1), ...,(x_D, y_D)\right \}$ as a bag. $x_d\ (d\in\{1,2,...,D\})$ are instances of $B$, each with labels $y_d \in {0,1}$. Consequently, the label $Y$ of $B$ is given by:
\begin{equation}
Y = \prod _{y_d\in\mathbf{y}}(y_d)
= \left\{\begin{matrix}
0, \forall y_d =0, \\ 
1, \exists y_d =1.
\end{matrix}\right.
\end{equation}

In this work, the features aggregation $\mathcal{A}(\cdot)$ is implemented by Attention-based multi-instances learning (AB-MIL), which integrates the strengths of attention-based MIL pooling for aggregating the features $\hat{\mathbf{Z}}^{pe}$ into a single feature vector $\mathbf{Z}^{pe}$ with a weighted averaging operation. Figure \ref{fusecpath}(c) demonstrates the procedure of feature aggregation via AB-MIL. In this work, the input of AB-MIL is the re-embedded patch-level FMs features $\hat{\mathbf{Z}}^{pe}\in \mathbb{R}^{(N_K\times K)\times D}$. By adopting AB-MIL as pooling module, the aggregated feature $\mathbf{Z}^{pe}$ can be formulated as follows:
\begin{equation}
\begin{aligned}
& \mathbf{Z}^{pe}=\mathcal{A}(\hat{\mathbf{Z}}^{pe})=\sum_{d=1}^{D}({a_d(\hat{\mathbf{Z}}^{pe}_d)\cdot \hat{\mathbf{Z}}^{pe}_d}), \\
& a_d(\hat{\mathbf{Z}}^{pe}_d)=\frac{ \exp\left ({\mathbf{W}^T \tanh \left (\mathbf{V} (\hat{\mathbf{Z}}^{pe}_d)^T\right )}\right)}{\sum_{j=1}^{D} \exp\left ({\mathbf{W}^T \tanh \left (\mathbf{V} (\hat{\mathbf{Z}}^{pe}_j)^T\right )}\right )}.
\end{aligned}
\end{equation}
where $a_d(\cdot)$ denotes the attention operation corresponding to the embedding $\hat{\mathbf{Z}}^{pe}_d$, $\mathbf{W} \in \mathbb{R}^{(N_K\times K)\times 1}$ and $V \in \mathbb{R}^{(N_K\times K)\times D}$ are learnable parameters. The aggregated feature $\mathbf{Z}^{pe}$ will be the input of the slide-level collaborative distillation module, which is an essential step to fuse the slide-level FMs.

\subsection{Slide-level collaborative distillation}
Slide-level foundation model is capable of yielding a high-level representation of WSI, which is more concise for a downstream task fine-tuning. However, it is a problem that fuse these slide-level global representations with patch-level local representations. This is challenging due to the dimensional gaps. In the proposed FuseCPath, we try to solve this problem by regarding the slide-level features as soft labels derived from teacher models. Consequently, we propose a slide-level collaborative distillation strategy to fuse slide-level FMs that contain global information simultaneously.

Consider the re-embedded patch-level features $\mathbf{Z}^{pe}$ and slide-level features $\mathbf{L}^1_{se},\ ..., \mathbf{L}^n_{se}$ derived from $N$ heterogeneous slide-level FMs $F^1_{se},\ ..., F^N_{se}$, each slide-level FM is regarded as a teacher model. We first project the embeddings $\mathbf{L}^1_{se},\ ..., \mathbf{L}^n_{se}$ with a linear layer to ensure the same dimensions of the features.
\begin{equation}
\mathbf{h}^i_{se}=\text{Linear}_i(\mathbf{L}^i_{se}),\mathbf{L}^i_{se}\in \mathbb{R}^{1\times d^i_{se}},
\label{diststeacher}
\end{equation}
where $\mathbf{h}^i_{se}$ denotes the projected features subject to the teacher model ${F}^i_{se}\in \mathcal{F}_{se}=\{\text{Gigapath-SE, TITAN, PRISM}\}$. $d^i_{se}$ denotes the dimension of slide-level embeddings, where $d^i_{se}\in \mathcal{D}_{se}=\{768,1280\}$. $\mathbf{h}^i_{se}$ is usually calculated by a Linear layer. Similarly, the patch-level FMs are regarded as a student model. The projection layer is formulated as follows.
\begin{equation}
\mathbf{h}_{pe}=\text{Linear}(\mathbf{Z}_{pe}),
\label{distsstu}
\end{equation}

To ensure performance, the features will be softened by temperature $\tau$ using the softmax function. In this work, the temperature $\tau$ is set to 3 for the classification task, and $\tau$ is set to 1 for the regression task.
\begin{equation}
\Bar{\mathbf{h}}^i_{se}=\text{SoftMax}(\frac{\mathbf{h}^i_{se}}{\tau}),\ \Bar{\mathbf{h}}_{pe}=\text{SoftMax}(\frac{\mathbf{h}_{pe}}{\tau}),
\label{soften}
\end{equation}

With the softened distribution $\Bar{\mathbf{h}}^i_{se}$ and $\Bar{\mathbf{h}}_{pe}$, the Kullback-Leibler Divergence $\text{KL}(\cdot\|\cdot)$ is adopted to formulate the distillation loss function $\mathcal{L}^i_{dist}$ of the teacher model (slide-level FM), which is formulated as follows.
\begin{equation}
\begin{aligned}
\mathcal{L}^i_{dist}& = \tau^2\cdot\text{KL}(\Bar{\mathbf{h}}_{pe}\|\Bar{\mathbf{h}}^i_{se}), \\
& = \tau^2\cdot \sum_{c=1}^C \Bar{h}^c_{pe}\log\frac{\Bar{h}^c_{pe}}{\Bar{h}^c_{se}},
\end{aligned}
\label{distsloss}
\end{equation}
where $C$ denotes the dimension of the softened distributions. Given label $\mathbf{y}$, the combined loss function of distillation is formulated as follows:
\begin{equation}
\mathcal{L}_{fuse}=\lambda \mathcal{L}_{task}(\mathbf{h}_{pe},\mathbf{y})+(1-\lambda)\frac{1}{N}\sum_{i=1}^N \mathcal{L}^i_{dist}.
\label{kdloss}
\end{equation}
where $\mathcal{L}_{task}$ is related to the downstream tasks. The $\mathcal{L}_{task}$ for biomarker prediction will be a binary cross-entropy loss function (mutation refers to 1, and wild-type refers to 0). For the prediction of gene expression, the task will be modeled as a regression problem. So $\mathcal{L}_{task}$ will be the Mean Squared Error (MSE). $\lambda$ denotes the weight for balancing the student and teacher models. For survival analysis, the $\mathcal{L}_{task}$ will be formulated using the Cox proportional hazard model \citep{10226279}. We summarize the main training procedure for our FuseCPath framework in Algorithm \ref{alg:fusecpath}.

\begin{algorithm}[!t]
\caption{Procedure of FuseCPath} 
\label{alg:fusecpath}
\SetKwInput{KwInput}{Input}    
\SetKwInput{KwOutput}{Output}  
  \KwInput{Whole Slide Image $X$; \\
  \qquad \quad Task-related label $\mathbf{y}$.}
  \KwOutput{Fused features $\mathbf{Z}^{f}$; \\
  \qquad \quad Prediction result $\hat{\mathbf{y}}$.}
  \SetKwFunction{FMain}{Main}
  \SetKwFunction{FNMSDe}{FuseCPath}
  \SetKwFunction{FSub}{Sub}
  \SetKwProg{Fn}{Procedure}{:}{}
  \Fn{\FNMSDe{X, $\mathbf{y}$}}{
    $\mathbf{H}^{f^{1}_{pe}},...,\mathbf{H}^{f^{n}_{pe}}\gets f^1_{pe}(X),...,f^n_{pe}(X)$; \\
    $\mathbf{L}^{f^{1}_{se}},...,\mathbf{L}^{f^{n}_{se}}\gets F^1_{se}(X),\ ..., F^N_{se}(X)$; \\
    $\mathbf{C}_1,...,\mathbf{C}_K\gets \text{Cluster}(\mathbf{H}^{f^{1}_{pe}},...,\mathbf{H}^{f^{n}_{pe}})$; \\
    $\mathbf{H}^{f^{1}_{pe}}_s,...,\mathbf{H}^{f^{N}_{pe}}_s\gets \text{Selection}(\mathbf{C}_1,...,\mathbf{C}_K)$; \\
    $\hat{\mathbf{Z}}^{pe}\gets \text{ReEmbedding}(\mathbf{H}^{f^{1}_{pe}}_s,...,\mathbf{H}^{f^{N}_{pe}}_s)$; \\
    $\mathbf{Z}^{pe}\gets \text{ReEmbeddedFeaturesAggregation}(\hat{\mathbf{Z}}^{pe})$; \\
    $\mathbf{Z}^{f}\gets \text{Distillation}(\mathbf{Z}^{pe};\mathbf{L}^{f^{1}_{se}},...,\mathbf{L}^{f^{n}_{se}})$; \\
    $\hat{\mathbf{y}}\gets \text{TaskHeader}(\mathbf{Z}^{f};\mathbf{y})$; \\
    \KwRet $\mathbf{Z}^{f}$, $\hat{\mathbf{y}}$; \\
  }
\end{algorithm}

\subsection{Implementation details}
The complete procedure of our FuseCPath framework consists of the following essential modules: multi-view patch features clustering, patch-level features re-embedding, features aggregation, and slide-level distillation. All experiments in this paper were finished on four NVIDIA A100 80G GPUs with an Ubuntu 20.04 system. The implementation of FuseCPath is mainly based on the \textit{Pytorch} framework, \textit{Trident} \citep{trident}, OpenSlide, and Scikit-Learn packages.

\textbf{Multi-view patch features clustering}. In the cluster module, the implementation mainly relies on the \textit{mvlearn} and \textit{Trident} packages. The input WSIs will first be fed into the Deeplabv3 model to extract the tissue regions. Then the patches are tiled from the tissue regions, and all the patches are sized by 256$\times$256. Patch embeddings are derived with patch-level FMs $\mathcal{F}_{pe}=$ \{CONCH, Virchow2, Gigapath\} using these tiled patches. Finally, the embeddings derived from distinct FMs will be concatenated into a list with the corresponding indices, which is the input of multi-view spectral clustering. The patches will be clustered into $K=50$ clusters.

\textbf{Patch-level features re-embedding}. In the re-embedding module, the implementation mainly relies on the $\text{R}^{2}$T and \textit{Trident} packages. We adopt hierarchical sparse self-attention ($\text{top-}k=8$) to accelerate the training process and meanwhile suppress the over-fitting problem \citep{10226279}. The input embeddings for training are selected from the clustered patch embeddings. We select $N_K=10$ patches from each cluster and a total of 500 patches for each WSI. The re-embedding contains two layers of \textbf{C-MSA} and one layer \textbf{CC-MSA} with 10\% dropout during training. For the detailed parameters for the re-embedding module, the batch size is 64. The learning rate starts with 1e-4. The model is optimized using Stochastic Gradient Descent (SGD), where the momentum parameter is $m=0.9$ and the learning rate decay ratio is 5e-5.

\textbf{Slide-level distillation}. In the slide-level distillation module, the implementation mainly relies on the \textit{PyTorch} and \textit{Trident} packages. We derive the slide embeddings from the slide-level FMs Gigapath-SE, TITAN, and PRISM for the soft labels during training. Each $\text{Linear}_i(\cdot)$ in Equation \ref{diststeacher} is implemented by a linear projection layer to align with the re-embedded patch-level features. To balance the weight for teacher models, we set $\lambda=0.5$ and $N=3$ in Equation \ref{kdloss} during training.

The repeated selection-based data augmentation plays a critical role in enhancing FuseCPath's performance. For 5-fold cross-validation, we partition all WSIs into training and validation sets with a ratio of 80\%:20\%. The repeated summarization process is applied separately to each partitioned dataset as follows: For each of $W$ WSIs, we first perform clustering to generate $K=50$ clusters, then randomly select $K_N=10$ patches from each cluster. This operation will generate 500 representative patches per WSI. By repeating this procedure $N_R=50$ times, we obtain $N_R=50$ distinct summarizations for each WSI, effectively expanding the dataset size from $W$ to $W\times N_R$. To address the remaining class imbalance, we apply conventional augmentation techniques, including random flipping, cropping, rotation, scaling, and blurring, to enhance the training dataset quality. The fused embeddings will be input to the Multi-layer perceptrons (MLPs) for prediction or regression tasks. 

\section{Experiment}
\subsection{Dataset description and evaluation metrics}
\begin{figure*}[!t]
\centerline{\includegraphics[width=\textwidth]{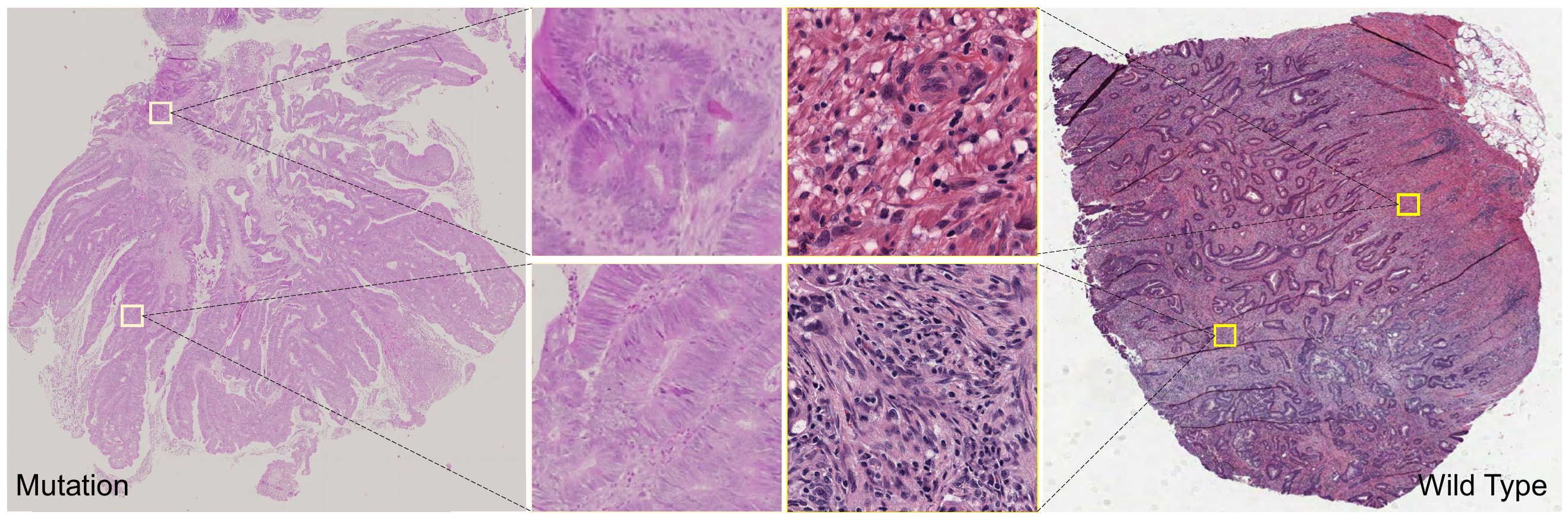}}
\caption{Examples of WSIs in TCGA-COAD dataset for mutation and wild type. Several patches are selected for visualization.}
\label{fig5}
\end{figure*}

\begin{table*}[ht]\footnotesize
\centering
\caption{The statistics of the WSIs corresponding to different biomarker mutations in TCGA-BLCA, TCGA-LUAD, and TCGA-COAD datasets.}
\label{table3}
\begin{tabular}{l|cc|ccccc|cc}
\hline
TCGA & \multicolumn{2}{c|}{BLCA} & \multicolumn{5}{c|}{LUAD} & \multicolumn{2}{c}{COAD} \\
Biomarkers  & TP53 & ATM & EGFR & FAT1 & KRAS & LRP1B & TP53 & BRAF & KRAS \\ \hline
Mutation    & 196  & 57  & 74  & 51    & 149  & 185 & 210 & 62   & 183  \\ 
Wild-Type & 210  & 349 & 483 & 506   & 408  & 371 & 347 & 366  & 245  \\ 
Total       & 406  & 406 & 557 & 557   & 557  & 557 & 557 & 428  & 428 \\ \hline
\end{tabular}
\end{table*}

To evaluate and compare the performance of the proposed FuseCPath framework with other baseline methods, we utilize the publicly available datasets in the Cancer Genome Atlas (TCGA) \citep{kandoth2013mutational} on three essential downstream tasks, which are biomarker prediction, gene expression prediction, and survival analysis. The statistics of the WSIs corresponding to different biomarker mutations in TCGA-BLCA, TCGA-LUAD, and TCGA-COAD datasets are summarized in Table \ref{table3}. Additionally, we present the examples of WSI and corresponding patches utilized in our datasets in Figure \ref{fig5}.

\textbf{TCGA-LUAD}. The lung adenocarcinoma cancer (LUAD) dataset contains 557 WSIs. 445 of them are selected as the training dataset, and 112 of them are selected as the validation dataset. The tasks of biomarker prediction for \textit{EGFR}, \textit{FAT1}, \textit{KRAS}, \textit{LRP1B}, and \textit{TP53} are utilized in our experiments. The corresponding survival times and censor state are provided for training.

\textbf{TCGA-BLCA}. The bladder cancer (BLCA) dataset contains 406 WSIs. 324 of them are selected as the training dataset, and 82 of them are selected as the validation dataset. The tasks of biomarker prediction for \textit{TP53} and \textit{ATM} are utilized in our experiments. The corresponding survival times and censor state are provided for training.

\textbf{TCGA-COAD}. The colon adenocarcinoma cancer (COAD) dataset contains 428 WSIs. 342 of them are selected as the training dataset, and 86 of them are selected as the validation dataset. The tasks of biomarker prediction for \textit{BRAF} and \textit{KRAS} are utilized in our experiments.

\textbf{Metrics for biomarker prediction}. The WSI-based biomarker prediction task can be modeled as a binary classification problem. Current studies typically evaluate the WSI-based classification methods using the Area Under the Receiver Operating Characteristic Curve (AUROC) metric. The AUROC is particularly suitable for the classification task. It provides a reliable assessment of classifier performance that accounts for both positive and negative sample classification across all decision thresholds, making it robust even with imbalanced data distributions.

\textbf{Metrics for gene expression prediction}. The WSI-based gene expression prediction can be modeled as a regression task. The prediction result is a vector containing each expression of the target gene. The prediction results ($\mathbf{y}_{pred}$) and ground truth ($\mathbf{y}$) are regarded as the input of MSE, which is formulated as follows:
\begin{equation}
\text{MSE}(\mathbf{y}_{pred},\mathbf{y})=\frac{1}{N} \sqrt{\sum^{N=10}_{i=1} \|\mathbf{y}^i_{pred}-\mathbf{y}^i\|^2},
\label{mse}
\end{equation}

\begin{table*}[!ht]\footnotesize
\centering
\caption{Comparisons of our proposed FuseCPath framework with different MIL-based methods to predict biomarkers on the TCGA-LUAD and TCGA-BLCA datasets. The \textbf{bold} results denote the highest scores, and the \underline{underlined} results denote the second-highest scores.}
\label{tablemilcomparison}
\begin{tabular}{l|cccc|cc|c}
\hline
AUROC & \multicolumn{4}{c|}{TCGA-LUAD} & \multicolumn{2}{c|}{TCGA-BLCA} & Average\\
Methods & \makecell[c]{EGFR} & \makecell[c]{FAT1} & \makecell[c]{KRAS} & \makecell[c]{LRP1B} & \makecell[c]{TP53} & \makecell[c]{ATM} \\
\hline
MeanMIL & 80.8$\pm$0.8 & 78.7$\pm$1.7 & 78.6$\pm$0.5 & 79.2$\pm$3.4 & 77.4$\pm$0.8 & 80.4$\pm$0.6 & 79.2$\pm$1.3 \\ 
MaxMIL & 80.1$\pm$1.2 & 79.3$\pm$1.4 & 80.1$\pm$1.0 & 80.0$\pm$0.6 & 78.0$\pm$1.1 & 79.9$\pm$1.0 & 79.6$\pm$1.1 \\ 
AB-MIL \citep{ilse2018attention} & 82.9$\pm$1.3 & 81.0$\pm$2.7 & 81.5$\pm$0.3 & 80.3$\pm$4.9 & 81.0$\pm$1.9 & 82.1$\pm$1.2 & 81.5$\pm$2.1 \\ 
TransMIL \citep{transmil} & 83.2$\pm$1.3 & 81.1$\pm$0.7 & 82.0$\pm$1.7 & 81.8$\pm$0.9 & 82.4$\pm$1.0 & 83.7$\pm$1.1 & 82.4$\pm$1.1 \\ 
R$^2$T-MIL \citep{tang2024feature} & \underline{86.4$\pm$1.2} & \underline{83.2$\pm$0.3} & \underline{84.9$\pm$0.8} & \underline{84.2$\pm$1.1} & \underline{84.5$\pm$0.8} & \underline{85.9$\pm$0.8} & \underline{84.9$\pm$0.8} \\ 
FuseCPath (\textbf{Ours}) & \textbf{89.5$\pm$0.7} & \textbf{85.8$\pm$1.2} & \textbf{86.8$\pm$0.1} & \textbf{86.4$\pm$1.0} & \textbf{86.0$\pm$1.1} & \textbf{88.3$\pm$0.6} & \textbf{87.1$\pm$0.8} \\ 
\hline
\end{tabular}
\end{table*}

\textbf{Metrics for survival analysis}. To evaluate the performance of survival analysis, we select the metric called the Concordance Index (C-index) for our comparisons. C-index measures the concordance of the ranking for predicted risk with the ground truth survival times, which is formulated as follows:
\begin{equation}
C_\text{index}=\frac{1}{n} \sum_{i\in \{i,...,n|\delta_i=1\}} \sum_{t_i>t_j} I\left[f_i>f_j\right].
\label{cindex}
\end{equation}
where $n$ denotes the number of pairs for comparisons. $I\left[\cdot\right]$ denotes the indicator function. $t$ denotes the observed survival time. $f$ denotes the corresponding predicted risk. The value of the C-index ranges from 0 to 1. A higher C-index presents a better survival prognosis and vice versa. When the C-index value is 0.5, the prediction is ineffective.

Another comparison metric is the univariate Kaplan-Meier survival curve with log-rank $p$-values. In survival analysis, the disease state changes over time. The Kaplan-Meier survival curve intuitively demonstrates the survival differences of patients in different groups, and the log-rank method can be further used to test the statistical significance of the differences.

\textbf{Metrics for clustering}. Due to the ground truth labels for the evaluation of clustering being unavailable, we select two widely adopted metrics for evaluating the clustering, which are the silhouette coefficient (SC) and the Calinski-Harabasz (CH) index. The value of SC ranges from -1 to 1, where values approaching 0 suggest cluster overlap, negative values indicate incorrect assignments of the clusters, and higher positive values reflect well-separated clusters. The CH index evaluates clustering quality by calculating the ratio of between-cluster variance to within-cluster variance, where variance is defined as the sum of squared Euclidean distances. Higher CH values indicate better clustering results, reflecting both strong separation between different clusters and high compactness within individual clusters.

\subsection{Biomarker predictions}
\textbf{Comparisons with MIL-based methods}. In this experiment, we perform a comprehensive evaluation of our FuseCPath framework against previous MIL-based methods, including MeanMIL, MaxMIL, AB-MIL, TransMIL, and vanilla R$^2$T-MIL. For a fair comparison, each MIL-based method is trained using fused foundation model features from CONCH \citep{conch}, Virchow \citep{virchow2}, and Gigapath \citep{gigapath} through a direct concatenation strategy.

To quantitatively evaluate the performance of each method, we present the results assessed by AUROC in Table \ref{tablemilcomparison}. The experimental results demonstrate that the proposed FuseCPath framework outperforms these baseline MIL-based methods. Performance improvements are observed across multiple biomarker prediction tasks on datasets TCGA-LUAD and TCGA-BLCA, with an average increase of 5\% in AUROC compared to the baseline methods with the best performance. The superior results can be attributed to the teacher model's high-level feature representations, which provide additional discriminative information to guide the student models' feature fusion process. These results prove that our re-embedding and distillation-based FuseCPath framework enhances feature learning, particularly in scenarios with class imbalance and limited labeled training data.
\begin{table}[!t]\footnotesize
\centering
\caption{Comparisons of our proposed FuseCPath framework with different embeddings from the FMs to predict biomarkers on the TCGA-LUAD and TCGA-COAD datasets. The \textbf{bold} results denote the highest scores, and the \underline{underlined} results denote the second-highest scores.}
\label{tablefmcomp}
\begin{tabular}{l|cc|c}
\hline
AUROC & \multicolumn{2}{c|}{TCGA-COAD} & \makecell[c]{Average} \\
Methods & \makecell[c]{BRAF} & \makecell[c]{KRAS} \\
\hline
CTransPath (\cite{ctranspath})        & 58.8$\pm$5.5 & 52.5$\pm$9.1 & 55.7$\pm$7.3 \\ 
Virchow (\cite{vorontsov2024virchow}) & 62.7$\pm$2.7 & 47.8$\pm$9.6 & 55.3$\pm$6.2 \\ 
CONCH (\cite{conch})                  & 59.4$\pm$3.0 & 55.3$\pm$6.1 & 57.4$\pm$4.6 \\ 
H-Optimus (\cite{hoptimus0})          & 84.7$\pm$0.7 & 49.6$\pm$5.7 & 67.2$\pm$3.2 \\ 
UNI (\cite{uni})                      & 73.4$\pm$3.1 & 56.7$\pm$4.5 & 65.1$\pm$3.8 \\ 
Gigapath (\cite{gigapath})            & 76.7$\pm$4.5 & \underline{61.4$\pm$8.1} & 69.1$\pm$6.3 \\ 
Virchow2 (\cite{virchow2})            & 83.0$\pm$2.6 & 60.9$\pm$1.8 & 71.9$\pm$2.2 \\ 
Gigapath-SE (\cite{gigapath})         & 50.0$\pm$5.1 & 51.8$\pm$4.4 & 51.0$\pm$4.9 \\ 
MADELEINE (\cite{jaume2024madeleine}) & 58.4$\pm$1.9 & 53.6$\pm$3.3 & 56.0$\pm$2.6 \\ 
CHIEF (\cite{chief})                  & 67.1$\pm$5.1 & 56.9$\pm$8.7 & 62.0$\pm$6.9 \\ 
PRISM (\cite{prism})                  & 57.2$\pm$1.9 & 57.1$\pm$7.6 & 57.2$\pm$3.8 \\ 
COBRA (\cite{cobra})                  & \underline{86.2$\pm$2.8} & 58.1$\pm$6.9 & \underline{72.3$\pm$4.9} \\ 
FuseCPath (\textbf{Ours})             & \textbf{91.8$\pm$3.0} & \textbf{78.1$\pm$5.4} & \textbf{84.9$\pm$4.2} \\ 
\hline
\end{tabular}
\end{table}

\begin{figure*}[h!]\centering
\centerline{\includegraphics[width=\textwidth]{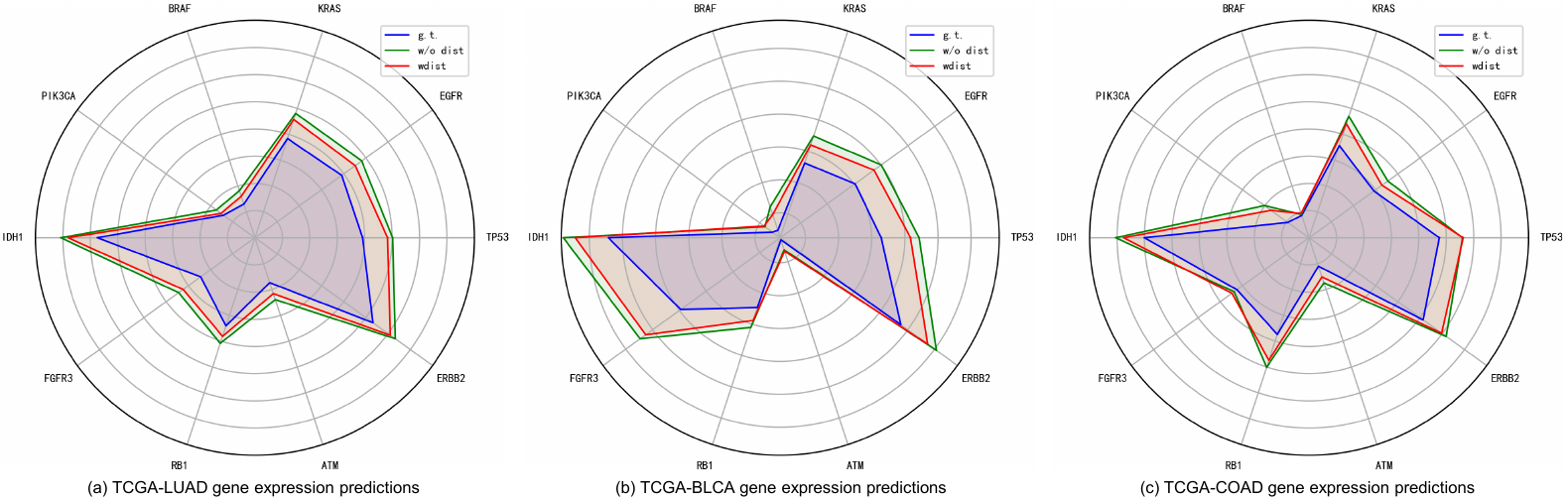}}
\caption{Visualized results of the gene expression predictions and the ground truth values observed by bulk RNA sequencing. If the values are closer to the ground truth, the prediction results are better.}
\label{fig:pathway}
\end{figure*} 

\begin{table*}[!ht]\footnotesize
\centering
\caption{The quantitative results of gene expression prediction. The expressions are calculated by Equation \ref{tpm}. If the values are closer to the ground truth, the prediction results are better.}
\label{table_rna}
\begin{tabular}{llllllllllll}
\hline
\multicolumn{2}{l}{\begin{tabular}[c]{@{}l@{}}Genetypes   /\\      Datasets\end{tabular}} & TP53  & EGFR  & KRAS  & BRAF  & PIK3CA & IDH1  & FGFR3 & RB1   & ATM   & ERBB2 \\
\hline
\multirow{3}{*}{TCGA-LUAD}                             & g.t.                             & 1.961 & 1.954 & 1.919 & 0.651 & 0.706  & 2.875 & 1.224 & 1.705 & 0.869 & 2.661 \\
                                                       & w/o dist                        & 2.310 & 2.205 & 2.206 & 0.711 & 0.668  & 3.342 & 1.520 & 1.844 & 0.997 & 2.961 \\
                                                       & wdist                            & 2.318 & 2.158 & 2.188 & 0.703 & 0.662  & 3.307 & 1.518 & 1.807 & 0.981 & 2.951 \\
\hline
\multirow{3}{*}{TCGA-BLCA}                             & g.t.                             & 2.144 & 2.016 & 1.814 & 0.739 & 0.766  & 3.214 & 2.475 & 1.736 & 0.652 & 2.869 \\
                                                       & w/o dist                        & 2.516 & 2.305 & 2.045 & 0.913 & 0.814  & 3.693 & 3.033 & 1.854 & 0.723 & 3.333 \\
                                                       & wdist                            & 2.486 & 2.270 & 2.004 & 0.890 & 0.822  & 3.612 & 3.029 & 1.842 & 0.735 & 3.270 \\
\hline
\multirow{3}{*}{TCGA-COAD}                             & g.t.                             & 2.375 & 1.462 & 1.783 & 0.432 & 0.478  & 3.004 & 1.625 & 1.870 & 0.551 & 2.571 \\
                                                       & w/o dist                        & 2.706 & 1.575 & 2.150 & 0.361 & 0.806  & 3.330 & 1.598 & 2.307 & 0.673 & 2.889 \\
                                                       & wdist                            & 2.709 & 1.542 & 2.105 & 0.369 & 0.759  & 3.281 & 1.637 & 2.273 & 0.658 & 2.887 \\
\hline
\end{tabular}
\end{table*}

\textbf{Comparisons with individual FMs}. In this experiment, we evaluate the classification performance of our proposed FuseCPath framework against state-of-the-art (SOTA) FMs across two biomarker prediction tasks \textit{BRAF} and \textit{KRAS} predictions in the TCGA-COAD dataset. To ensure a fair comparison, we have reproduced all baseline methods using their embeddings with the same classifier implementation. 

The comprehensive evaluation results are presented in Table \ref{tablefmcomp}, which is measured by AUROC, reveal several key findings: First, FuseCPath consistently outperforms all individual FMs across the prediction tasks on TCGA-COAD. The observed average performance has improved by 17\%. This improvement can be attributed to two fundamental advantages of our FuseCPath framework: First, the effective fusion of complementary features from heterogeneous FMs through our proposed re-embedding and distillation mechanism. Second, the patch-level and slide-level simultaneous fusion of multiple FMs adaptively emphasizes the most meaningful features for the prediction of each biomarker. These results imply that an effective fusion of diverse FMs can yield superior predictive capability compared to a single model, as the ensemble approach mitigates individual model limitations while preserving their respective strengths through feature fusion. The results also prove the importance of the fusion of pathology FMs, demonstrating that a carefully devised fusion framework can improve the model's performance by leveraging the rich but complementary information contained in diverse FMs.

\subsection{Gene expression prediction}
In this experiment, we evaluate the performance of gene expression prediction for our method. The proposed FuseCPath is capable of predicting the expression of many genes involved in pathways. We select 10 popularly investigated genes to assess the prediction errors of the proposed FuseCPath, which are \textit{TP53, EGFR, KRAS, BRAF, PIK3CA, IDH1, FGFR3, RB1, ATM,} and \textit{ERBB2}. Expressions are evaluated by logarithmically transformed transcripts per million (TPM) values $t$, which are formulated as follows:
\begin{equation}
t = \log_2(\text{TPM}+1).
\label{tpm}
\end{equation}
We select two methods for our comparisons, which are the proposed FuseCPath (wdist) and the method only containing patch-level features using R$^2$T without the distillation module (w/o dist). Figure \ref{fig:pathway} presents radar charts comparing predicted and ground truth gene expression, visually illustrating the alignment between our model's predictions and ground truths. We present the quantitative results averaged across all samples from the validation datasets of TCGA-LUAD, TCGA-BLCA, and TCGA-COAD in Table \ref{table_rna}, respectively. To evaluate the prediction error, we provide the mean square error (MSE) comparisons between the prediction results of different methods and the ground truth in Table \ref{msepred}, demonstrating the accuracy of the prediction for individual genes.

From these results, we can conclude that FuseCPath effectively predicts gene expression using the embeddings containing enough knowledge distilled from multiple FMs, without requiring additional specialized knowledge from genomics training data. Additionally, the average prediction error of the complete model of FuseCPath remains below 30\% for all genes in this experiment, indicating consistent performance across different genetic targets and types of cancers. The findings indicate that the distillation mechanism enables more efficient utilization of useful information contained in multiple FMs.
\begin{table}[!t]\footnotesize
\centering
\caption{Performance comparison on the gene expression prediction with different methods, which are evaluated with Mean Squared Error (MSE). The bold results denote the best scores. Lower values are closer to the ground truth.}
\label{msepred}
\begin{tabular}{lccc}
\hline
Methods     & TCGA-LUAD           & TCGA-BLCA            & TCGA-COAD            \\
\hline
w/o dist    & 0.265          & 0.329           & 0.280           \\
wdist       & \textbf{0.250} & \textbf{0.298}  & \textbf{0.256}  \\
\hline
\end{tabular}
\end{table}

\subsection{Survival Analysis}
\begin{figure*}[!ht]\centering
\centerline{\includegraphics[width=\textwidth]{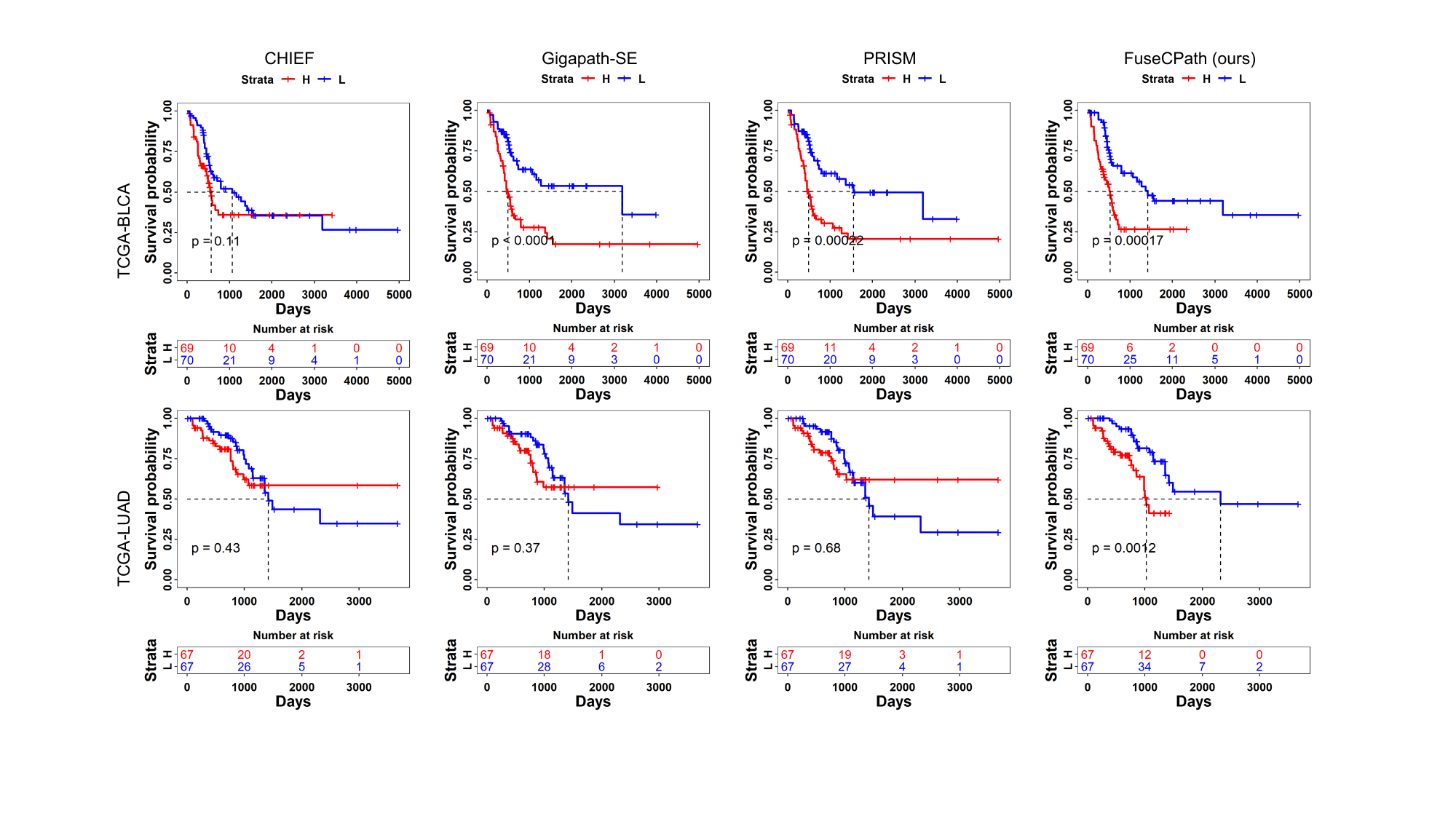}}
\caption{Kaplan-Meier survival curves of FuseCPath and representative slide-level FMs on TCGA-BLCA and TCGA-LUAD datasets. }
\label{fig:kmcurve}
\end{figure*}

In this experiment, we evaluate and compare the performance of the proposed FuseCPath with the features from several slide-level FMs, which are CHIEF \citep{chief}, Gigapath-SE \citep{gigapath}, and PRISM \citep{prism}. We provide the results of Kaplan-Meier survival curves for each comparison method in Figure \ref{fig:kmcurve}. The test cohorts are divided into high- and low-risk groups using the median risk score predicted by our proposed FuseCPath framework. Comparative analysis demonstrates that FuseCPath achieves significantly improved risk stratification, producing a more distinct separation between the two risk groups with enhanced prognostic discrimination capability. This implies that the FuseCPath consistently performs better than the other FMs in distinguishing high- and low-risk patients.

\begin{table}[!t]\footnotesize
\centering
\caption{Performance comparison on the survival analysis with different slide-level foundation model features, which are evaluated with C-index. The bold results denote the highest scores and the \underline{underlined} results denote the second-highest scores.}
\label{cindextab}
\begin{tabular}{lcc}
\hline
Methods     & TCGA-BLCA           & TCGA-LUAD      \\
\hline
CHIEF    & 0.614$\pm$0.037          & 0.644$\pm$0.052   \\
Gigapath-SE    & \underline{0.649$\pm$0.039}    & \underline{0.659$\pm$0.050}   \\
PRISM    & 0.629$\pm$0.036              & 0.634$\pm$0.046   \\
FuseCPath (\textbf{Ours})       & \textbf{0.706$\pm$0.031} & \textbf{0.708$\pm$0.049}  \\
\hline
\end{tabular}
\end{table}

To further evaluate the performance of FuseCPath, we conduct quantitative experiments using the metric C-index, and the results are presented in Table \ref{cindextab}. Compared with the state-of-the-art slide-level FMs, we can find that the performance of FuseCPath is the highest value in C-index among all comparison methods. The C-index is improved by 8.8\% and 7.4\% on the dataset TCGA-BLCA and TCGA-LUAD over the second-best method, respectively. The prediction performance of the proposed FuseCPath is better than that of individual FMs, which implies that the model will benefit from the knowledge from both patch-level and slide-level embeddings.

\subsection{Analysis of clustering}
Before the training process of the FuseCPath, one essential step is to select representative image patches from the input WSIs. In this work, to integrate the features from heterogeneous patch-level FMs, we devise a multi-view clustering-based strategy to partition the patches into $K=50$ clusters. Each embedding from the corresponding foundation model can be regarded as a view of the WSI. In this section, we conduct an experimental analysis of clustering.
\begin{figure*}[!ht]
\centerline{\includegraphics[width=0.96\textwidth]{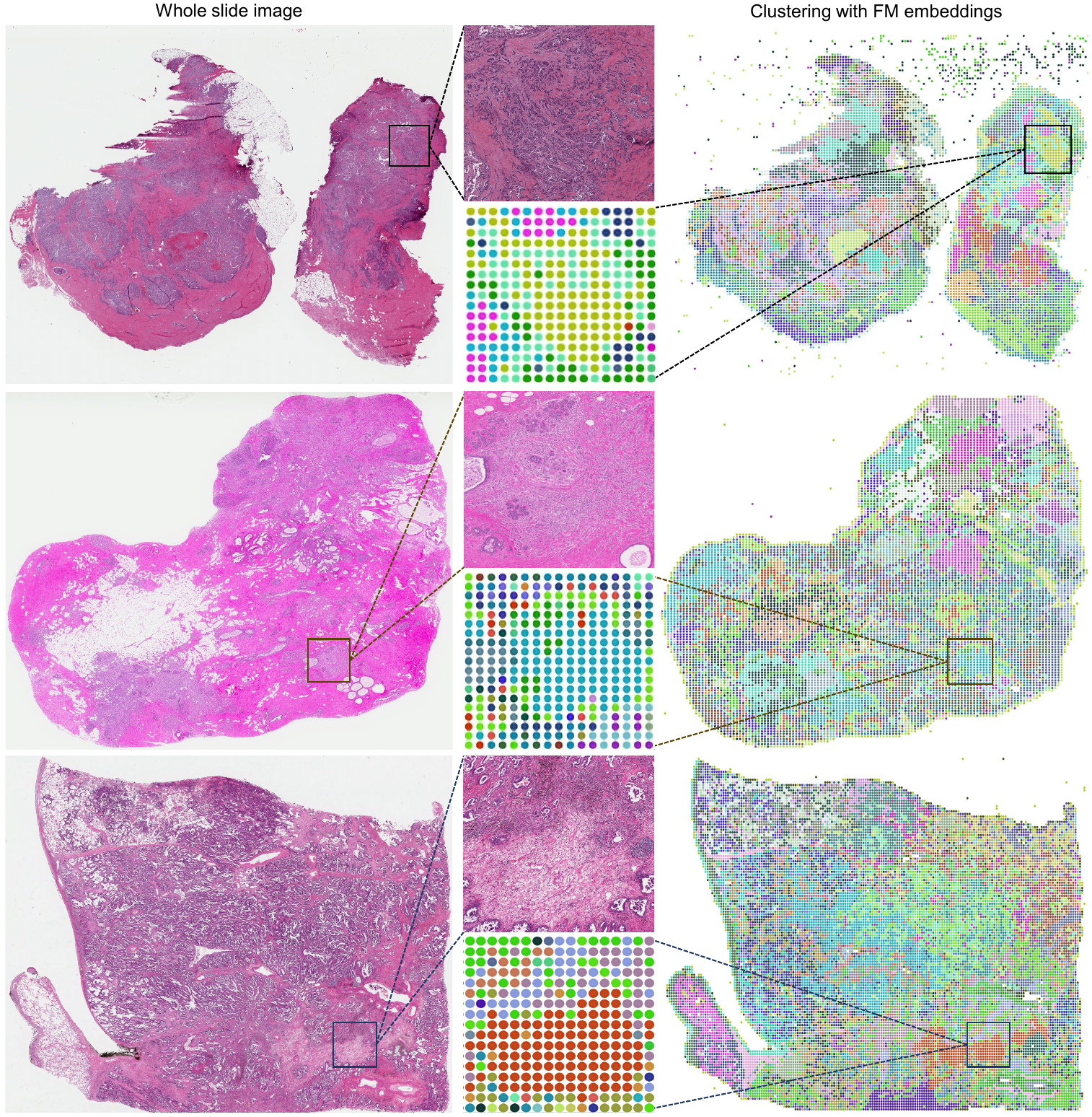}}
\caption{Visualized results of patch multi-view clustering ($K=50$) based on patch embeddings derived from heterogeneous FMs.}
\label{fig:cluster}
\end{figure*}

\textbf{Visualization and interpretability}. We present the visualized results of the clustering for each example WSI in Figure \ref{fig:cluster}. To better demonstrate the visualization results, we present the zoom-in areas in original WSIs alongside their corresponding clustering results. The results clearly show that under the guidance of embeddings from the heterogeneous FMs, the clustering results exhibit clear alignment with cellular morphological distributions. The clustered regions are related to tissue structures, indicating that the multi-view clustering can capture both local and global morphological patterns. This implies that the patches selected from clusters are representative and reliable enough for the training of FuseCPath.
\begin{table}[!t]\footnotesize
\centering
\caption{Performance comparison on the multi-view clustering with different numbers of clusters (\# Clusters), which are evaluated with SC and CH.}
\label{clsk}
\begin{tabular}{lcc}
\hline
\# Clusters & SC            & CH              \\ \hline
$K$=30    & 0.09          & 832.5           \\
$K$=40    & 0.10          & 946.1           \\
$K$=50    & \textbf{0.15} & \textbf{1025.6} \\
$K$=60    & 0.13          & 997.6          \\
$K$=70    & 0.11          & 879.4           \\ \hline
\end{tabular}
\end{table}

\begin{table}[!t]\footnotesize
\centering
\caption{Performance comparison of clustering methods, which are evaluated with SC and CH.}
\label{cls}
\begin{tabular}{lcc}
\hline
Clustering   methods        & SC   & CH  \\ \hline
Spectral   Clustering      & 0.07                     & 209.1                     \\
Agglomerative   Clustering & 0.07                     & 893.3                     \\
Affinity   Propagation     & 0.09                     & 713.8                     \\ 
Multi-view Clustering                    & \textbf{0.15}                     & \textbf{1025.6}                    \\ \hline
\end{tabular}
\end{table}
\textbf{Analysis of the number of clusters}. To determine the optimal number of clusters ($K$) for our multi-view clustering, we performed a systematic evaluation guided by both quantitative metrics and biological considerations. To ensure an adequate representation of histological patterns, we set a lower bound of 30 clusters in this experiment. As fewer clusters probably affect the diversity of selected patches. We validate this range by evaluating the quality of the cluster at 10-cluster intervals in Table \ref{clsk}. The selected metrics are the silhouette coefficient (SC) and Calinski-Harabasz (CH). From the results, we can conclude that $K=50$ emerges as the optimal choice that satisfies both our computational metrics and the visual constraints of biological tissues. So we ultimately selected $K=50$ clusters for the training of FuseCPath. 

\textbf{Multi-view clustering \textit{vs.} single-view clustering}. In this part, we devise experiments to compare multi-view clustering with other single-view clustering. We select spectral clustering, agglomerative clustering, and affinity propagation for comparisons. These selected methods are clustered with embeddings from CONCH \citep{conch}.  We present SC and CH quantitative results in Table \ref{cls}. The results prove that multi-view clustering achieves a higher quality of clustering compared to the other single-view clustering methods. Multi-view clustering integrates complementary features from heterogeneous patch-level FMs into a unified representation space, whereas single-view clustering, such as spectral clustering, only operates on a single feature space. Consequently, multi-view clustering will capture higher-level relationships between different feature spaces, enabling nonlinear pattern discovery beyond single-view clustering limitations.

\subsection{Ablation studies}
To evaluate the effectiveness of the main components of our proposed FuseCPath framework, we conduct ablation studies on the following aspects: the number of slide-level FMs and the number of selected patches. All experiments were conducted on the prediction task of biomarker \textit{TP53} for TCGA-LUAD and TCGA-BLCA datasets.
\begin{table}[!t]\footnotesize
\centering
\caption{Ablation study on the number of teacher models (Slide-level FMs) to predict biomarker \textit{TP53} on the TCGA-LUAD and TCGA-BLCA datasets. FuseCPath$^{+}$ indicates that the slide-level distillation module is eliminated from the complete framework. G denotes Gigapath-SE. P denotes PRISM. T denotes TITAN.}
\label{tableslidecomparison}
\begin{tabular}{lccc}
\hline
AUROC & \makecell[c]{LUAD} & \makecell[c]{BLCA} & Average \\
Methods & \makecell[c]{TP53} & \makecell[c]{TP53} & \\
\hline
FuseCPath$^{+}$ (0FM) & 85.7 & 83.0 & 84.4 \\ 
FuseCPath$^{+}$+G (1FM) & 86.5 & 83.6 & 85.1 \\ 
FuseCPath$^{+}$+G+P (2FMs) & 87.2 & 85.2 & 86.2 \\ 
FuseCPath$^{+}$+G+P+T (3FMs) & \textbf{89.5} & \textbf{86.0} & \textbf{87.8} \\ 
\hline
\end{tabular}
\end{table}

\begin{table}[!t]\footnotesize
\centering
\caption{Ablation study on the number of patches for training to predict biomarker \textit{TP53} on the TCGA-LUAD and TCGA-BLCA datasets.}
\label{patches}
\begin{tabular}{lccc}
\hline
AUROC & \makecell[c]{LUAD} & \makecell[c]{BLCA} & Average \\
\# Patches & \makecell[c]{TP53} & \makecell[c]{TP53} & \\
\hline
$N_K$=300 & 86.7 & 83.2 & 85.0 \\ 
$N_K$=400 & 87.8 & 85.7 & 86.8 \\ 
$N_K$=500 & \textbf{89.5} & \textbf{86.0} & \textbf{87.8} \\ 
$N_K$=600 & 88.6 & 84.7 & 86.9 \\ 
$N_K$=700 & 88.1 & 85.5 & 86.8 \\ 
\hline
\end{tabular}
\end{table}

\begin{figure}[!ht]
\centerline{\includegraphics[width=\columnwidth]{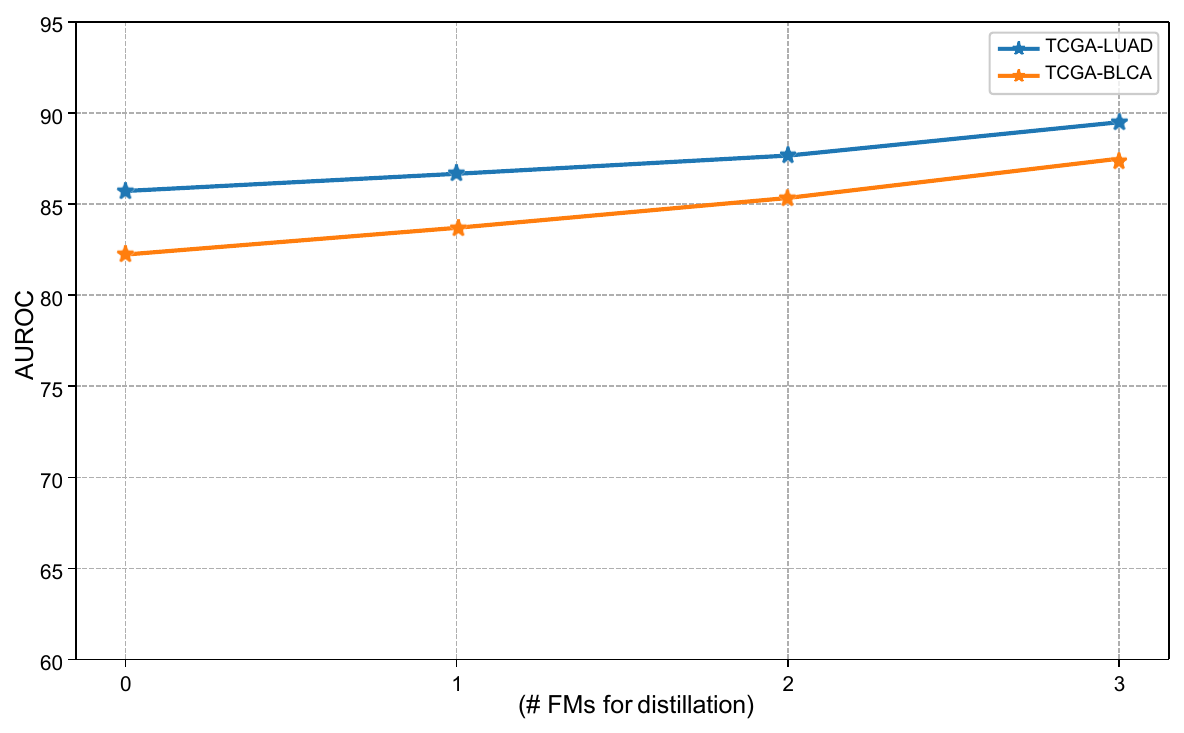}}
\caption{Ablation study on the number of teacher models (Slide-level FMs). The AUROC is increasing with the number of teacher models.}
\label{fig:nfms}
\end{figure}

\begin{figure}[!ht]
\centerline{\includegraphics[width=\columnwidth]{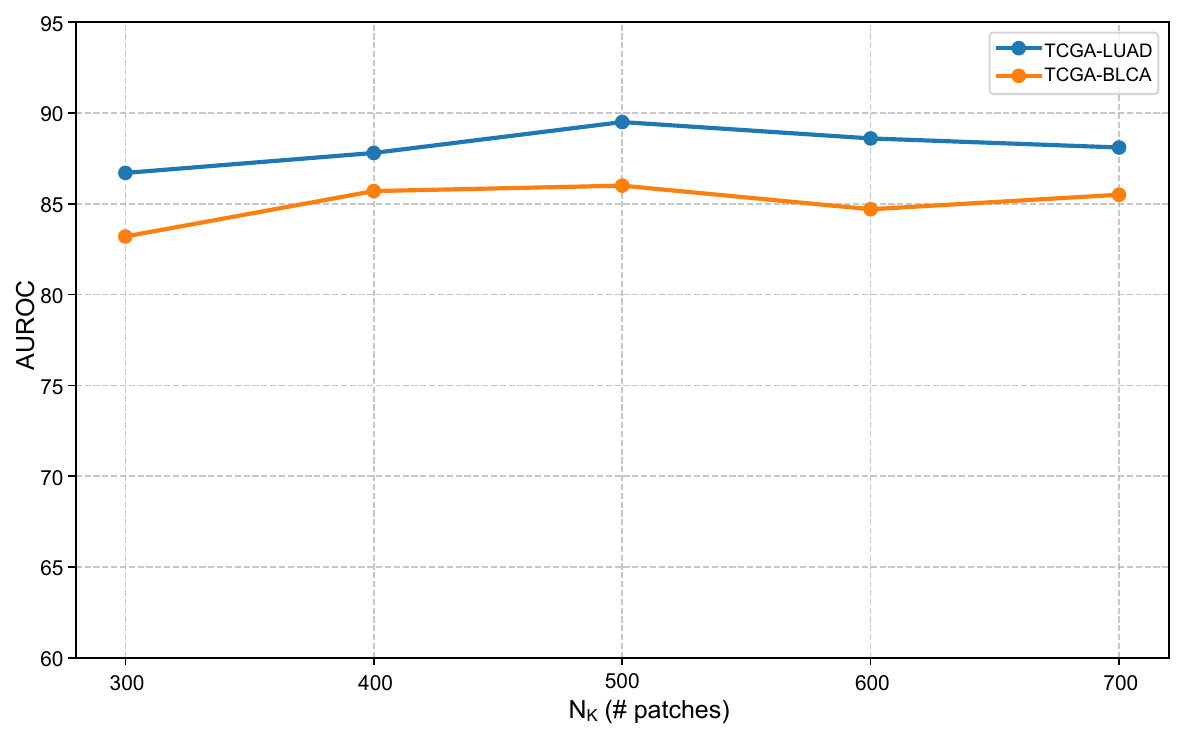}}
\caption{Ablation studies on the number of selected patches ($N_K$) for training. The best performance measured by AUROC is observed by $N_K$=500.}
\label{fig:patches}
\end{figure}
\textbf{Ablation studies on the number of slide-level FMs}. In this experiment, we conduct an ablation study on the number of slide-level FMs for distillation.  We present the quantitative results of AUROC in Table \ref{tableslidecomparison} and Figure \ref{fig:nfms}. The best prediction performance (AUROC) is obtained when 3 slide-level FMs are utilized. The performance is decreasing when the number of slide-level FMs decreases. When FuseCPath is only trained with re-embedded patch-level features, the performance decreases by 3.6\% on average. These results imply that more slide-level FMs utilized for distillation will provide more useful semantic information during training.

\textbf{Ablation studies on the number of selected patches}. In this experiment, we conduct an ablation study on the number of selected patches $N_K$ for patch-level features re-embedding during training. We present the quantitative results of AUROC in Table \ref{patches} and Figure \ref{fig:patches}. The best prediction performance (AUROC) is obtained when $N_K$=500, which means that 10 patches are selected from 50 clusters in total. When $N_K < $500, the performance will increase as more patches are selected for training, this is because more patches will summarize the semantic information of WSIs more comprehensively. When $N_K > $500,  the performance will slightly decrease because of the overfitting problem. Consequently, we select $N_K$ = 500 to implement the proposed FuseCPath framework.

\section{Discussion and limitation}
In this work, we have proposed a novel framework called FuseCPath for the fusion of multi-scale heterogeneous pathology FMs simultaneously. The proposed FuseCPath framework includes the following essential modules to effectively fuse the pathology FMs: representative patches selection based on multi-view patch features clustering, patch-level features re-embedding \& aggregation, and slide-level collaborative distillation. These modules contribute to the performance improvement of biomarker prediction, gene expression, and survival analysis on datasets TCGA-LUAD, TCGA-BLCA, and TCGA-COAD. In conclusion, the FuseCPath framework will yield a new ensemble model with superior performance and benefit many meaningful downstream tasks. Additionally, clustering with multi-view features will provide insight into the visualization analysis of tissue morphography in WSI analysis. 

Despite the demonstrated utility in this article, the FuseCPath poses potential limitations in its capacity to integrate more high-dimensional multi-omics data, such as the emerging spatial transcriptomic technologies. The current framework may not fully capture the underlying nonlinear relationships between different omics data. The rapid evolution of FMs presents a promising strategy for the integration of multi-omics data and WSIs \citep{omiclip}. In future research, we will extend the FuseCPath framework to the fusion of multi-omics representations and embeddings from multi-omics FMs to improve the precision of molecular-level WSI analysis.

\section*{Declaration of Competing Interest}
The authors declare that they have no known competing financial interests or personal relationships that could have appeared to influence the work reported in this paper.

\section*{CRediT authorship contribution statement}
\textbf{Zhidong Yang}: Methodology, Investigation, Writing - original draft \& editing. \textbf{Xiuhui Shi}: Conceptualization, Resources, Data curation. \textbf{Wei Ba}: Conceptualization, Data curation. \textbf{Zhigang Song}: Conceptualization, Data curation. \textbf{Haijing Luan}: Methodology, Validation. \textbf{Taiyuan Hu}: Methodology, Validation. \textbf{Senlin Lin}: Conceptualization, Validation. \textbf{Jiguang Wang}: Resources, Writing - review \& editing. \textbf{Shaohua Kevin Zhou}: Resources, Writing - review \& editing. \textbf{Rui Yan}: Methodology, Resources, Writing - review \& editing. 

\section*{Acknowledgments}
This study was funded by the National Natural Science Foundation of China (62402473, 62271465), Beijing Natural Science Foundation (L252175), and the Suzhou Basic Research Program (SYG202338). J.W. lab is supported by RGC grant (R6003–22), ITC grant (ITCPD/17-9), Padma Harilela Professorship, and Sinovac Fellowship Program.





\bibliography{sn-bibliography}

\end{document}